\PassOptionsToPackage{pdfpagelabels=false}{hyperref} 

\documentclass[10pt,journal,compsoc]{joser1}

\usepackage{url}

\ifCLASSINFOpdf
   \usepackage[pdftex]{graphicx}
\else
   \usepackage[dvips]{graphicx}
\fi

\journalnumber{1}                       
\journalvolume{1}                       
\journalmonth{April}                
\journalyear{2014}                      
\articlefirstpage{1}                  
\articlelastpage{10}                   
\copyrightauthor{D. Coleman, I. A. \c{S}ucan, S. Chitta, N. Correll}
\headoddname{D. COLEMAN et al./ Reducing the Barrier to Entry of Complex Robotic
Software}%

\begin{document}
\title{Reducing the Barrier to Entry of Complex \\\vskip 0.3\baselineskip
Robotic Software: a MoveIt! Case Study}

\author{
David COLEMAN$^{1}$
\qquad
Ioan \c{S}UCAN$^{2}$
\qquad
Sachin CHITTA$^{3}$
\qquad
Nikolaus CORRELL$^{1}$


\IEEEcompsocitemizethanks{\IEEEcompsocthanksitem Parts of this work were
performed while D. Coleman, I. A. \c{S}ucan, and S. Chitta were at Willow
Garage, Inc. D. Coleman and N. Correll are supported by NASA grant
NNX12AQ47G.\protect\\

\IEEEcompsocthanksitem Authors retain copyright to their papers and grant JOSER
unlimited rights to publish the paper electronically and in hard copy. Use of
the article is permitted as long as the author(s) and the journal are properly
acknowledged.}

} 

\address{
$^1$ Dept. of Computer Science, University of Colorado at Boulder, 430 UCB,
Boulder, CO 80309\\
$^2$ Google Inc., 1600 Amphitheatre Parkway, Mountain View, CA 94043\\
$^3$ SRI International, 333 Ravenswood Avenue, Menlo Park, CA 94025
}

\markboth

\IEEEcompsoctitleabstractindextext{%
\begin{abstract}
Developing robot agnostic software frameworks involves synthesizing the
disparate fields of robotic theory and software engineering while simultaneously
accounting for a large variability in hardware designs and control paradigms. As
the capabilities of robotic software frameworks increase, the setup difficulty
and learning curve for new users also increase. If the entry barriers for
configuring and using the software on robots is too high, even the most powerful
of frameworks are useless. A growing need exists in robotic software engineering
to aid users in getting started with, and customizing, the software framework as
necessary for particular robotic applications. In this paper a case study is
presented for the best practices found for lowering the barrier of entry in the
MoveIt! framework, an open-source tool for mobile manipulation in ROS, that
allows users to 1) quickly get basic motion planning functionality with minimal
initial setup, 2) automate its configuration and optimization, and 3) easily 
customize its components. A graphical interface that assists the user in
configuring MoveIt! is the cornerstone of our approach, coupled with the use of
an existing standardized robot model for input, automatically generated
robot-specific configuration files, and a plugin-based architecture for
extensibility. These best practices are summarized into a set of
\textit{barrier to entry design principles} applicable to other robotic
software. The approaches for lowering the entry barrier are evaluated by usage
statistics, a user survey, and compared against our design objectives for their
effectiveness to users.
\end{abstract}

}

\maketitle

\section{Introduction}
\IEEEPARstart{M}{anaging} {the increasing complexity of modern robotic software
is a difficult engineering challenge faced by roboticists today. The size of the
code bases of common open source robotic software frameworks such as ROS
\cite{quigley2009ros}, MoveIt! \cite{moveit} and OROCOS
\cite{bruyninckx2001open} continues to increase \cite{makarenko2007benefits},
and the required breadth of knowledge for understanding the deep stack of
software from control drivers to high level planners is becoming more daunting.
As it is often beyond the capabilities of any one user to have the necessary
domain knowledge for every aspect of a robot's tool chain, it is becoming
increasingly necessary to assist users in the configuration, customization, and
optimization of the various software components of a reusable robotic
framework. 

The user interface design principles required in the emerging field of robotics
software is similar to other more mature software engineering fields and much
can be learned from them. There have been many examples of software, such as
computer operating systems, that have historically required many installation
and configuration steps whose setup process has since improved. Still, the user 
interface design principles for robotics is unique in 1) the degree to which software interacts with
hardware and real world environments compared to consumer-level software, 2) the large variety in complexity and
scale of robotic platforms, and 3) the long term desire to increase the autonomy
of robotics systems by reducing reliance on GUIs and increasing high level
robotic intelligence. 

\subsection{Barriers to Entry} 

The term \textit{barriers to entry} is used in the context of robotic software
engineering to refer to the time, effort, and knowledge that a new user must
invest in the integration of a software component with an arbitrary robot. This
can include, for example, creating a virtual model of the robot's geometry and
dynamics, customizing configuration files, choosing the fastest algorithmic
approach for a specific application, and finding the best parameters for various
algorithms.

Powerful robotics software generally requires many varying degrees of
customization and optimization for any particular robot to operate properly.
Choosing the right parameters for each utilized algorithm, software component,
and application typically involves expert human input using domain-specific
knowledge. Many new users to a software package, particularly as robotics
becomes more mainstream, will not have the breadth of knowledge to customize
every aspect of the tool chain. When the knowledge of a new user is insufficient
for the requirements of the software, the barriers to entry become
insurmountable and the software unusable. One of the emerging requirements of
robot agnostic frameworks is implementing mechanisms that will automatically
setup and tune task pipelines for arbitrary robots.

Another motivation for lowering the barrier to entry of complex robotics
software is the \textit{paradox of the active user}. This paradox explains a
common observation in many user studies that \textit{users never read manuals}
but start attempting to use the software immediately
\cite{carroll1987interfacing}. The users's desire to quickly accomplish a task
results in their skipping the reading of any provided documentation or gaining
deeper understanding of the system and instead diving right into completing
their task. The \textit{paradox} is that the users would actually save time in
the long run if they learned more about the system before attempting to use it,
but these studies showed that in reality people do not tend to invest time
upfront into learning a new system.

Even experts in the area of the associated robotics software will become
frustrated with robotics software if all initial attempts to setup and configure
the framework fail and no progress is made. Most researchers and engineers
typically do not have the time or ability to completely understand the entirety
of robotics software before they start using it. It is important for the
user's initial experience with a piece of software to be positive to ensure its
continued use.

\subsection{Benefits of Larger User Base}

The need to lower the barrier of entry is beneficial to the software itself in
that it enables more users to utilize the framework. If the software framework
is being sold for profit, the benefits of a larger user base are obvious. If
instead the software is a free open-source project, as many successful robotic
frameworks currently are \cite{makarenko2007benefits}, lowering the barrier to
entry is very beneficial in that it creates the \textit{critical mass of skilled
contributors} that has been shown to make open source projects successful
\cite{bruyninckx2001open}. As the number of users increases, the speed in which
bugs are identified and fixed increases \cite{schmidt1999software}. It is also
typically hoped that development contributions to the code base increases,
though this correlation is not as strong \cite{schmidt1999software}. One of the
key strengths of a larger community for an open source project is increased
participation of users assisting with quality assurance, documentation, and
support 
\cite{schmidt2001leveraging}.

Another benefit of lowering the barrier of entry is that it allows the robotics
software to become an educational tool for robotics. Not only is the software
accessible for academic research and industrial applications, but graduate,
undergraduate, and even primary-level students can use it to learn some of the
higher level concepts of robotic applications as has been demonstrated in
\cite{correll2013one, moll2011teaching, guyot2011teaching}. 

Beyond the motivation of success for an individual software project, broadening
access to robotics software development increases the number of creative minds
working on solving today's challenging robotics problems. Making the
accessibility of robotic development more like mobile device development and web
development might increase the speed of innovation in robotics, similar to that
experienced by phone apps and the Internet \cite{boudreau2012let}.

As used in this paper, the target \textit{users} are engineers, scientists,
students, and hobbyists with a general aptitude for software and robotics, but
who are not necessarily experts in either of those fields. The hope remains that
human-robotic interaction for the general population in the future will be based
on more natural methods and that software configuration and graphical user
interfaces (GUIs) are only necessary for the robot developers themselves
\cite{yancotaxonomy}.

\subsection{Difficulty of Robot Agnostic Software}

The software engineering challenges faced in making reusable, robot agnostic code are hard and are different than those in other re-usable software frameworks. In \cite{smart2007common}, Smart argues there are three main factors that make general-purpose robotic software difficult: heterogeneity of robotics, limited resources (computational or otherwise), and the high rate of hardware and software failures. The variety of different tasks and task constraints imposed upon robots is another challenge for robot agnostic software \cite{kchir2013top}.

The heterogeneity of robots is of primary concern to us in this paper -- accounting for different types of actuators, sensors, and overall form factors is a very difficult task. To some users a robot is a robust and precise industrial arm, to others a robot is simply a mobile base with wheels and a computer, and to others a robot is a fully anthropomorphic biped. Creating reasonable abstractions for these large amounts of variation requires many trade offs to be made that almost always lead to a sub-optimal solution for all robots. It is more difficult to create a hardware abstraction for a robot than a standard computer -- when robotic software must interact with physical devices through an abstraction, it gives up specific knowledge of the hardware that then requires much greater reasoning and understanding of its configuration \cite{smart2007common}.

Operational requirements are another challenge in making reusable software for a wide range of robotic platforms. Some users require hard real-time constraints, while others can tolerate ``fast enough'' or ``best effort'' levels of performance. Variable amounts of available computational resources such as processing power or the ``embeddedness'' of the system also makes it difficult to design robot agnostic code that can run sufficiently on all robots. The amount of required error checking and fault tolerance varies by application area, for example, there are significant differences between a university research robot and a space exploration rover or a surgical robot.	

In making robotic agnostic software, many time-saving shortcuts employed for single-robot software must be avoided. This includes hard coding domain-specific values for ``tweaking'' performance and using short-cutting heuristics applicable to only one hardware configuration. Instead reasonable default values or automatically optimized parameters must be provided as discussed later. 

On top of these challenges, packaging reusable software into an easy to setup experience for end users requires creating tools that automate the configuration of the software.

\subsection{Related Work}

There has been much work to address the software engineering challenges of
complex robotic frameworks, but typically the identified design goals have
emphasized the need for features such as platform independence, scalability,
real-time performance, software reuse, and distributed layouts
\cite{realtime_framework, collett2005player, kramer2007development}. In
\cite{kramer2007development}'s survey of nine open source robotic development
environments, a collection of metrics was used which included documentation and
GUIs, but no mention was made of setup time, barrier to entry, or automated
configuration.

A focus on component-based design of motion planning libraries similar to MoveIt! was addressed in \cite{brugali2010component}. The challenges of software reuse, combining various algorithms, and customizations are discussed, but the work falls short of addressing the initial ease of use of these robotics frameworks. The difficulty of creating good component abstractions between hardware and algorithms is addressed in \cite{kchir2013top}.

The importance of an open source robotics framework having a large number of
researchers and engineers motivated to contribute code and documentation is
emphasized in the OROCOS framework \cite{bruyninckx2001open}, which we
emphatically agree with, but take a step further by creating additional tools to
encourage higher user adoption.
  
Human-robot interaction (HRI) has also been a popular area of research, but
HRI's focus has been on the runtime behavior of robots and not on the
difficulties of human users applying software frameworks to robot hardware
\cite{hci_metrics, yancotaxonomy, goodrichseven}. For example, in
\cite{rescueRobots}, an effective user interface is presented for teleoperation
of rescue robots, but no thought is given to making it robot agnostic or to its
configuration. 

In \cite{chitta2012perception}, Chitta et. al. we presented a set of tools that
allowed the Arm Navigation software framework (the precursor to MoveIt!) to be
easily configured within a short amount of time for a new robotic system. This
paper extends and improves that work, focusing specifically on the difficulties
of setting up and configuring robotics software.

\subsection{Contribution and Outline}

In this paper, we will present best practice principles for lowering the barrier
of entry to robotic software using the new MoveIt! software \cite{moveit} as our
case study. In section \ref{sec::motion_planning} we will motivate the many
software components that make software like MoveIt! a good example of the
difficulties of complex robotics software. In section \ref{sec::moveit}, we will
briefly describe MoveIt! itself. In section \ref{sec::requirements} we explain
the design principles used to address the user interface needs of robotics
software. We then in section \ref{sec::lowering_barriers} show how we have taken these principles and implemented them
in an entry tool for MoveIt! to reduce the entry barriers. In section \ref{sec::results}, we will present the
results of these implementations on the size of the user base and ease of
adoption of the MoveIt! software framework. We will discuss our experiences and
lessons learned in section \ref{sec::discussion}, followed by our conclusion in
section \ref{sec::conclusion}.

\section{Motion Planning Frameworks}
\label{sec::motion_planning}

Robotic motion planning is a maturing and central field in robotics
\cite{moll2011teaching} that turns a high level task command into a series of
discrete motions such that a robot can move within its environment. The typical
use case considered in this paper is the problem of controlling a robotic arm
from one configuration to another while taking into account various constraints.

The software development of a motion planning framework (MPF) is challenging and
involves combining many disparate fields of robotics and software engineering
\cite{perez2010roadmap}. We refer to the software as a \textit{framework} in
this context because it abstracts the various components of motion planning into
generic interfaces as discussed later.

One of the most important features of a MPF is providing the structures and
classes to share common data between the different components. These basic data
structures include a model of the robot, a method for maintaining a
representation of the state of the robot during planning and execution, and a
method for maintaining the environment as perceived by the robot's sensors (the
``planning scene'').

In addition to the common data structures, a MPF requires many different
interacting software components, henceforth referred to as the \textit{planning
components}. A high level diagram of the various planning components is shown in
Figure \ref{fig:motionplanning_highlevel}. The planning component that actually
performs motion planning includes one or more algorithms suited for solving the
expected problems a robot will encounter. The field of motion planning is large,
and no one-size-fits-all solution exists yet, so a framework that is robot
agnostic should likely include an assortment of algorithms and algorithm
variants.

\begin{figure}[!t]
\centering
\includegraphics[width=3.4in]{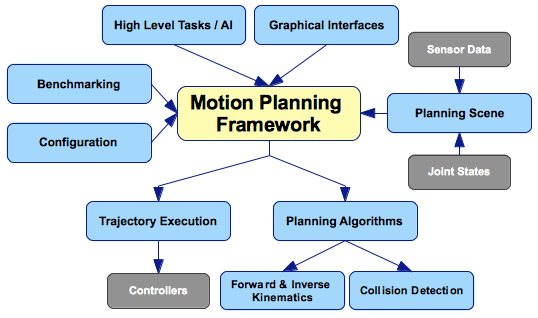}
\caption{High level diagram of various planning components (blue boxes) in a Motion Planning
Framework (MPF). Grey boxes represent external input and output.}
\label{fig:motionplanning_highlevel}
\end{figure} 

Other planning components include a collision checking module that detects the
potential intersection of geometric primitives and meshes in the planning scene
and robot model. A forward kinematics solver is required to propagate the
robot's geometry based on its joint positions, and an inverse kinematics solver
is required when planning in the Cartesian space of the end effector for some planning techniques. Other potential constraints, such as
joint/velocity/torque limits and stability requirements, require additional
components.

Secondary components must also be integrated into a powerful MPF. Depending on
what configuration space a problem is solved in, the generated motion planning
solution of position waypoints must be parameterized into a time-variant
trajectory to be executed. A controller manager must decide the proper low level
controllers for the necessary joints for each trajectory. A perception interface
updates the planning scene with recognized objects from a perception pipeline as
well as optional raw sensor data.

Higher level applications are built on top of these motion planning components
to coordinate more complex tasks, such as pick and place routines. Other
optional components of a MPF can include benchmarking tools, introspection and
debugging tools, as well as the user-facing GUI.

\subsection{Existing Motion Planning Software}
\label{sec::existing}

Many open source software projects for motion planning exists whose intent is to
provide a platform for testing and developing novel path planning algorithms and
other motion planning components. We will distinguish them from a motion
planning \textit{framework} due to their exclusion of actual hardware perception
and control. All offer varying degrees of modularity and all have a basic
visualization window for viewing motion plans of 3D geometries. A brief review
of them is presented here.

Both LaValle's Motion Strategy Library (MSL) \cite{lavallemsl}, 2000, and
Latombe's Motion Planning Kit (MPK) \cite{mpk}, 2003, have scopes limited to
only simulation and therefore are not frameworks in our definition. The MSL is
configured manually using six required text files and up to fifteen optional
files, depending on the planning problem. It has a GUI for tweaking parameters
and controlling the visualization of plans. The MPK is able to load robots with
varying geometry without recompiling code and provides a scene format that is an
extension of the Open Inventor format. It does not have a fully interactive GUI but
rather allows control only through keyboard shortcuts. Neither MSL or MPK
provides assistance for setting up a new robot and has little to no
documentation on this process.

The Karvaki Lab's Object-Oriented Programming System for Motion Planning
(OOPSMP) \cite{oopsmp}, 2008, is a predecessor to the Open Motion Planning
Library (OMPL) \cite{sucan2012the-open-motion-planning-library}, 2010, both of
which are collections of planning algorithms and components whose scope also
excludes hardware execution and perception tasks. OOPSMP is XML based for
configuration, scene definitions, and robot geometry. An additional SketchUp
interface provides a quick way to build environments. It has some GUIs that
assist in visualization. OMPL differs in its handling of environments and robots
in that it abstracts that notion into a black box and instead operates in
various configuration spaces. Neither OOPSMP nor OMPL provides any tools or GUIs
for configuration.

Diankov's OpenRave \cite{diankov2008openrave}, 2010, is a fully featured motion
planning framework with many high level capabilities, some GUIs, and the ability
to connect to hardware controllers and sensors. It uses the Collada format
\cite{collada}, as well as its own proprietary format, to define robots and
environments. Its main interface is through simple python scripting, and it
utilizes a plugin interface to provide extensibility of the framework. It too
falls short of providing easy to setup tools for new robots. 

Willow Garage's ROS Arm Navigation framework \cite{chitta2012perception}, 2010,
is the predecessor of MoveIt! and provides much of the same functionality of
MoveIt! and OpenRave but also includes a Setup Wizard that provides a GUI for
helping new users setup arbitrary robots into the framework. It was the
inspiration for the Setup Assistant described later in this paper.
\subsection{MoveIt! Motion Planning Framework}
\label{sec::moveit}

MoveIt!\cite{moveit} is the primary software framework for motion planning and
mobile manipulation in ROS and has been successfully integrated with many robots
including the PR2 \cite{wyrobek2008towards}, Robonaut
\cite{ambrose2000robonaut}, and DARPA's Atlas robot. MoveIt! is written entirely
in C++ but also includes Python bindings for higher level scripting. It follows
the principle of software reuse as advocated for robotics in
\cite{makarenko2007benefits} of not tying itself exclusively to one robotic
framework---in its case ROS---by creating a formal separation between core
functionality and robotic framework-dependent aspects (e.g., communication
between components).

MoveIt! uses by default the core ROS build and messaging systems. To be able to easily swap
components MoveIt! uses plugins for most of its functionality: motion planning
plugins (currently using OMPL), collision detection (currently using the Fast
Collision Library (FCL) \cite{fcl}), kinematics plugins (currently using  the
OROCOS Kinematics and Dynamics Library (KDL) \cite{kdl} for forward and inverse
kinematics for generic arms as well as custom plugins). The ability to change
these default planning components is discussed in section
\ref{subsec:extensiblity}. MoveIt!'s target application is manipulation (and
mobile manipulation) in industrial, commercial and research environments. For a
more detailed description of MoveIt!, the interested reader is referred to
\cite{moveit}.

\section{Entry Barrier Design Principles}
\label{sec::requirements}

In designing the configuration process that enables MoveIt! to work with many
different types of robots, with almost any combination of planning components,
several contending design principles for lowering the barrier of entry emerged.
These requirements were drawn partially from standard HCI principles
\cite{galitz2007essential}, from work on MoveIt!'s predecessor, and from an
iterative design process where feedback was gained from internal users at Willow
Garage during development. We believe these \textit{entry barrier design
principles} transcend motion planning and can be applied to most robotic
software:

{\bf Immediate}: The amount of time required to accomplish the most primitive
task expected from the robotic software component should be minimized. This is
similar to the time-honored ``Hello World'' demo frequently used by programming
languages and typical Quick Start guides in documentation. Immediacy is
essential for the \textit{paradox of the active user} as it provides cursory
feedback to the user that the software works and is worth investing further
time.

{\bf Transparent}: The configuration steps being performed automatically for the
user, and the underlying mechanisms utilized in the software components, should
be as visible as possible. Transparency is important so that users can later
understand what parameters are specific to their robot and know how to customize
the aspects they desire. A ``layered'' approach of presenting information can
offer a good balance of separating the required knowledge for a user's immediate
goals from the ``useful later'' information needed to prevent the user from
being hindered in the future.

{\bf Intuitive}: The need to read accompanied documentation, and the amount of
required documentation, should be minimized. A well-designed user interface, be
it graphical or command line, should be as intuitive as possible by following
standard design patterns and providing interface context clues. An ideal GUI for
configuration would not require any documentation for most users.

{\bf Reconfigurable}: The automatically generated parameters and default values
for the initial setup of a robot should be easy for the user to modify at a
later time. Typically, these parameters and values are chosen to work for the
largest number of robots possible but are not optimal for any particular robot.
Providing easy methods to reconfigure the initial setup is important for
allowing better performance. 

{\bf Extensible}: The user should be enabled to customize as many components and
behaviors as possible within the reasonable scope of the software. Providing the
means to extend the software with custom solutions for a particular application
makes the software far more powerful and re-usable for varying use-cases. A
typical solution for this is providing a plugin interface.

{\bf Documented}: The amount of reference material explaining how to use the
software should be maximized for as many aspects and user levels as possible.
Even the most intuitive software requires documentation for various aspects of
the operation or modification of the software itself. Different types of
documentation are needed for different users---for example developers and end
users---though in Robotics these groups are frequently the same. Documentation
is arguably the most important factor in reducing the barrier to entry of new
software \cite{forward2002relevance}.

These principles are additionally applicable to computer software in general, but a greater focus on hardware variance and the needs of developers has been applied. \textit{Reconfigurability}, or personalization, is common in computer software as well, but in our application we use it mainly in reference to parameters that require customization for different physical geometries and hardware designs. Similarly, \textit{extensibility} and \textit{transparentness} are design principles that aid robotic developers in applying their software to specific hardware. Whereas today most computer hardware is fairly standardized and share similar capabilities, robotics hardware still has large variability in design and capability. For this reason, \textit{transparency} is particularly important since many robotic researchers and developers need to understand the software enough to adapt it to their unique hardware.

Many of these \textit{entry barrier design principles} have opposing objectives that require a balance to be
found between them. For example, the desire for \textit{transparency} in the
underlying mechanisms often leads to slower setup times (lack of
\textit{immediacy}) and more complicated configuration steps (lack of
\textit{intuitiveness}). The need for extensibility of various components in the
software often results in far more complicated software design as more
abstraction is required, resulting in a less \textit{intuitive} code base and
difficult \textit{documentation}. Nevertheless, compromises can be made between
these principles that result in a superior user experience, as will be
demonstrated in the next section.

\section{Methods to Lower The Entry Barrier}
\label{sec::lowering_barriers}

One of the unique features of MoveIt! is the ratio of its power and features to
the required setup time. A beginner to motion planning can take a model of their
robot and with very little effort execute motion plans in a virtual environment.
With a few additional steps of setting up the correct hardware interfaces, one
can then execute the motion plans on actual robotic hardware.

The \textit{entry barrier design principles} discussed above were applied to
MoveIt! to address the challenges faced for new users to this complex software
framework. Developing these solutions required difficult software challenges to
be overcome as discussed in the following case study.

\subsection{Basic Motion Planning Out of the Box}

To address the entry barrier design principle of \textit{immediacy}, a
streamlined ``Quick Start'' for MoveIt! was created that consists of a series of
fairly trivial steps, relative to our target users. The most challenging of
these steps---creating a \textit{robot model}---is not directly related to the
configuration of MoveIt! but rather is a prerequisite of using the software
framework. Nevertheless, we will discuss this important prerequisite before
proceeding to the more directly-related configuration steps.

{\bf Robot Model Format}: The robot model is the data structures and
accompanying file format used to describe the three-dimensional geometric
representation of a robot, its kinematics, as well as other properties relevant
to robotics. These other properties can include the geometric visualization
meshes, courser-grained collision geometry of the robot used for fast collision
checking, joint limits, sensors, and dynamic properties such as mass, moments of
inertia, and velocity limits. Often the robot's joints and links relationships
are represented by a kinematic tree, though this approach is problematic when a
robot has a closed chain. In our application, as well as most state of the art
MPFs, we will restrict our definition of modeled robots to arbitrarily
articulated rigid bodies. 

\textit{Extensible} robotics software requires using a standardized format that
can express the intricacies of varying hardware configurations. An additional
design requirement for this standardized format is that it is \textit{intuitive}
for users to setup. There are a few options for representing robots, and in
MoveIt! it was accomplished by using the Unified Robotic Description Format (URDF \cite{urdf}) Document Object Model. This data structure is populated by reading human-readable (\textit{transparent}) XML schemas -- both URDF-formatted files (different from the datastructure) as well as the industry standard Collada \cite{collada} format. 

Creating an accurate model of a robot can be a difficult task. URDF models for
many robots already exist, so often users can avoid this problem. However, when
a custom robot requires a new robot model, the URDF model in ROS was found to be
the most appropriate to use since the user can also take advantage of tools in
ROS for working with the URDF. In particular, there are tools for verifying the
validity of the XML, for visualizing it, and for converting a SolidWorks CAD
model of a robot directly into URDF format. 

{\bf MoveIt! Setup Assistant}: The main facility that provides out of the box
support for beginners is the MoveIt! Setup Assistant (SA). The SA is a GUI that
steps new users though the initial configuration requirements of using a custom
robot with the motion planning framework (Figure \ref{fig:setupassistant}). It
accomplishes the objective of \textit{immediacy} for the user by automatically
generating the many configuration files necessary for the initial operation of
MoveIt!. These configurations include a self-collision matrix, planning group
definitions, robot poses, end effector semantics, virtual joints list, and
passive joints list. 

\begin{figure}[!t]
\centering
\includegraphics[width=3.4in]{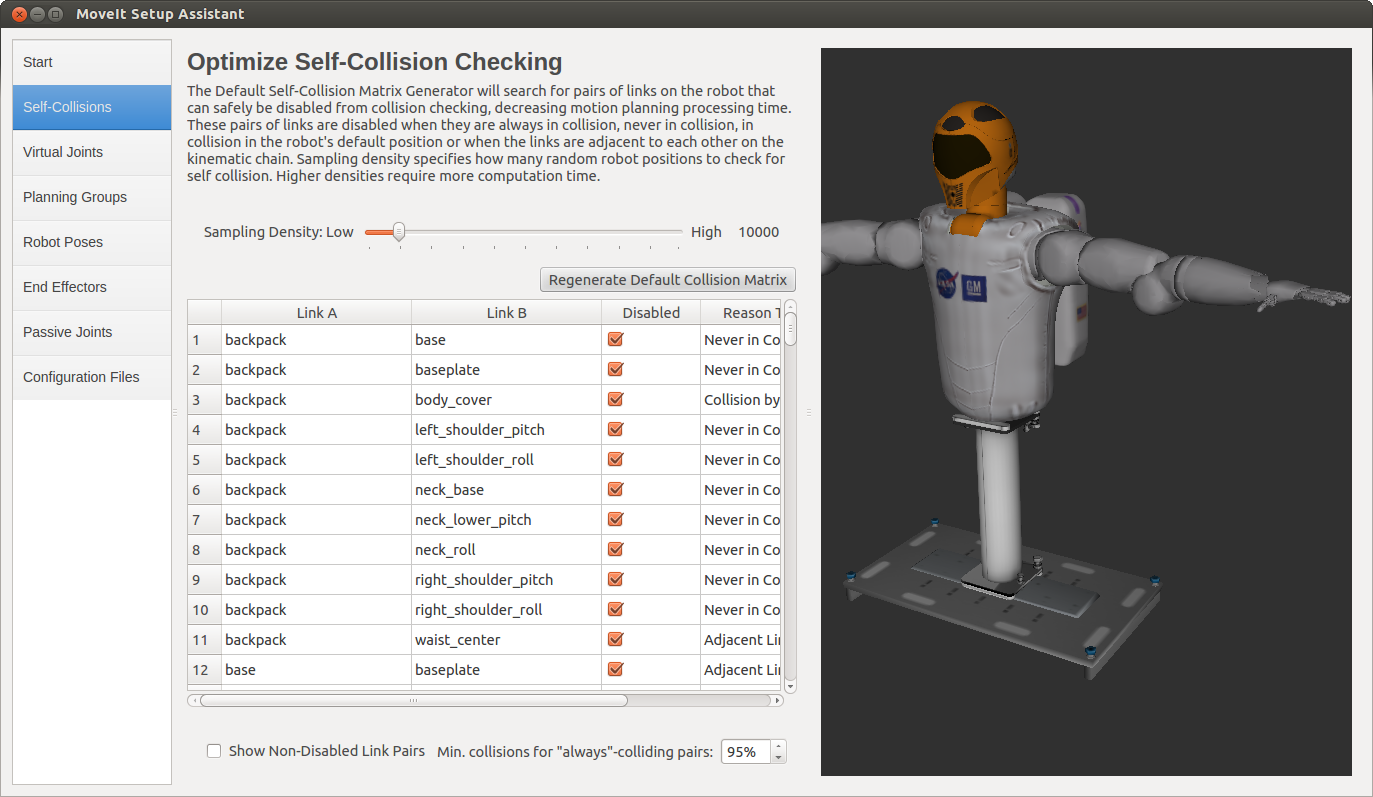}
\caption{MoveIt! Setup Assistant GUI with the NASA Robonaut loaded on the
self-collision matrix screen.}
\label{fig:setupassistant}
\end{figure}

The GUI consists of 1) a large navigation pane on the left that allows the user
to move back and forth through the setup process as needed (providing quick
\textit{reconfigurability}), 2) the middle settings window that changes based on
the current setup step being performed by the user, and 3) a right side
visualization of the three dimensional model of the robot as it is being
configured. The right side visualization increases the \textit{immediacy} of
results and \textit{transparency} of the configuration by highlighting various
links of the robot during configuration to visually confirm the actions of the
user, as shown in Figure \ref{fig:setupassistant3}.

\begin{figure}[!t]
\centering
\includegraphics[width=3.4in]{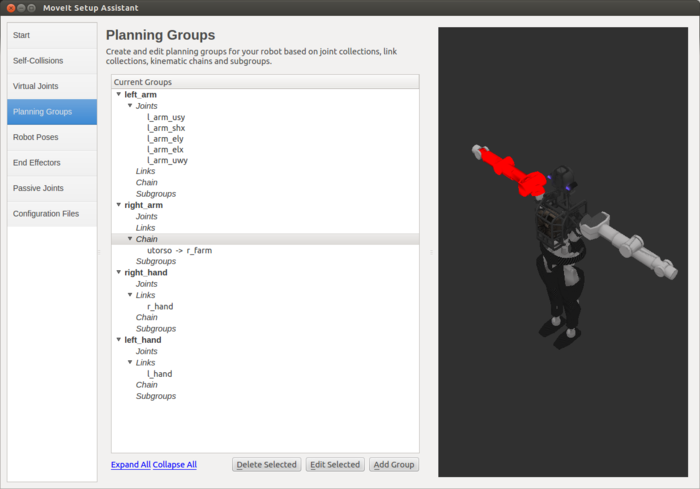}
\caption{MoveIt! Setup Assistant GUI with the Atlas robot's left arm highlighted
for user feedback on the planning groups screen.}
\label{fig:setupassistant3}	
\end{figure}

Using a properly formatted robot model file with the SA, MoveIt! can
automatically accomplish many of the required tasks in a MPF. If one desired,
the steps within the SA could almost entirely be automated themselves, but they
have been kept manual so to 1) increase \textit{transparency} and 2) provide
\textit{extensibility} for edge cases and unusual customizations. For example, automated semantical guesses of where an arm ends and an end effector begins can sometimes be incorrect.

{\bf MoveIt! Motion Planning Visualization GUI}: The details of the automated
configuration are left for the next section, but after the steps in the SA are
completed a demo script is created that automatically starts up a visualization
tool with the new robot loaded and ready to run motion planning algorithms in a
non-physics based simulation. A typical demo task would be using the computer
mouse to visually drag 3D interactive arrows situated on the robot's end
effector from a start position to a goal position around some virtual obstacle.
The demo can then quickly plan the arm in a collision free path around the
obstacle and visualize the results within the GUI. 

This user interaction is accomplished with the MoveIt! Motion Planning
Visualization (MMPV) \cite{moveit}, an additional GUI that allows beginning
users to learn and experiment with a large subset of the functionality provided
by MoveIt! (Figure \ref{fig:motionrvizplugin}). While the long term goal of
robotics is to provide more autonomous solutions to motion planning and
human-robot interactions \cite{yancotaxonomy}, the MMPV fulfills the immediate
needs of direct operation for testing and debugging the framework's capabilities
easily. This interface is a vital component of MoveIt!'s strategy to provide
\textit{immediate} results for motion planning with a robot that does not
require any custom coding. Once the user is comfortable with the basic feature
set and functionality of MoveIt!, \textit{extensibility} is provided via varying
levels of code APIs for more direct, non-GUI, access to the robot's abilities.

\begin{figure}[!t]
\centering
\includegraphics[width=3.4in]{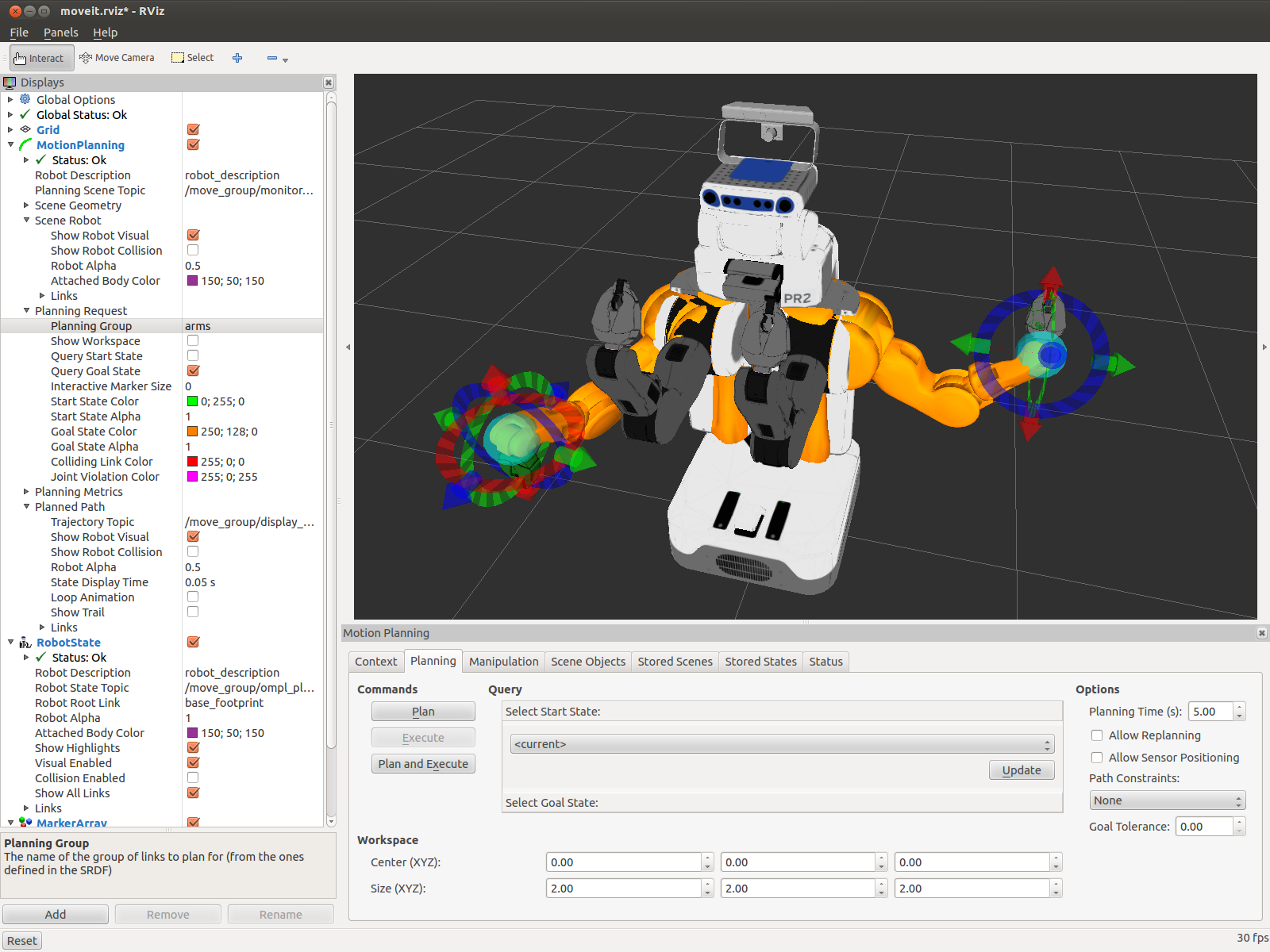}
\caption{MoveIt! Motion Planning Visualization GUI with the PR2 planning with
both arms to goal positions with interactive mouse-based tools}
\label{fig:motionrvizplugin}
\end{figure}

\begin{figure}[!t]
\centering
\includegraphics[width=3.4in]{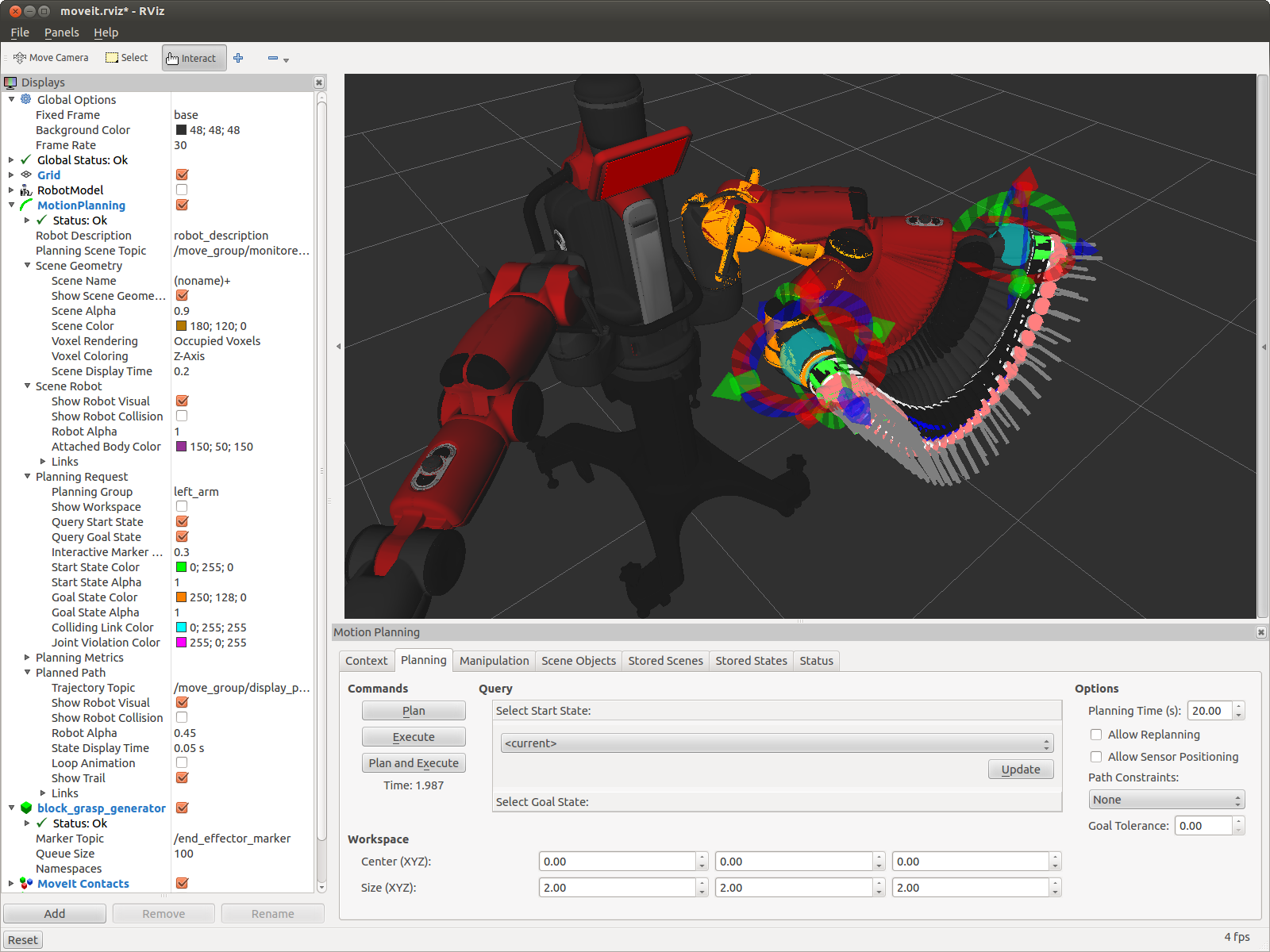}
\caption{MoveIt! Motion Planning Visualization GUI with the Baxter robot
visualizing steps of a motion plan}
\label{fig:motion_plan}	
\end{figure}

The MMPV provides a large number of features and visual tools for motion
planning. Using the MMPV, visualizations such as seen in Figure
\ref{fig:motion_plan} are provided of:

\begin{itemize}
    \item Start and goal configurations of the robot for planning
    \item Current robot hardware configuration
    \item Animated planned path before execution
    \item Detected collisions
    \item Sensor data and recognized objects
    \item Pick and place data such as grasp positions
    \item Attached bodies such as manipulated objects
    \item Planning metrics
\end{itemize}

Additionally, the MMPV contains many other non-visualization tools such as:
\begin{itemize}
    \item Connecting to a database of planning scenes
    \item Adjusting inverse kinematic settings
    \item Changing the utilized planning algorithm
    \item Adjusting the workspace size
    \item Adjusting goal tolerance and planning time
    \item Tweaking manipulation plans
    \item Loading and moving collision objects
    \item Exporting/importing scenes and states
    \item Viewing the status of MoveIt!
\end{itemize}

{\bf Hardware Configuration and Execution}: Once the user is comfortable with
the basic tools and features provided by MoveIt!, the next step is to configure
their robot's actual hardware actuators and control interfaces to accept
trajectory commands from MoveIt!. This step is not as easy and requires some
custom coding to account for the specifics of the robot hardware---the
communication bus, real-time requirements, and controller implementations. At
the abstract level, all MoveIt! requires is that the robot hardware exposes its joint positions and accepts a
standard ROS trajectory message containing a discretized set of time-variant
waypoints including desired positions, velocities, and accelerations.  

\subsection{Automatic Configuration and Optimization}

The size and complexity of a feature-rich MPF like MoveIt! requires many
parameters and configurations of the software be automatically setup and tuned
to improve the MPF's \textit{immediacy}. MoveIt! accomplishes this in the 1)
setup phase of a new robot, using the Setup Assistant, 2) during the runtime of
the application, and 3) using benchmarking and parameter
sweeping\cite{cohen2012generic}.

{\bf Self-Collision Matrix}: The first step of the SA is the generation of a
self-collision matrix for the robot that is used in all future planning to speed
up collision checking. This collision matrix encodes pairs of links on a robot
that never need to be checked for self-collision due to the kinematic
infeasibility of there actually being a collision. Reasons for disabled
collision checking between two links includes:
\begin{itemize}
    \item Links that can never intersect due to the reachability kinematics of
the robot
    \item Adjacent links that are connected and so are by design in collision
    \item Links that are always in collision for any other reason, including
inaccuracies in the robot model and precision errors
\end{itemize}

This self-collision matrix is generated by running the robot through tens of
thousands of random joint configurations and recording statistics of each link
pair's collision frequency. The algorithm then creates a list of link pairs that
have been determined to never need to be collision checked. This reduces future
motion planning runtimes because it reduces the amount of required collision
checks for every motion planning problem. The algorithm is incomplete because in
probabilistically rare cases a pair of links will be disabled for collision
checking when they should not be. For this reason, the number of tests needs to
be very high.

{\bf Configuration Files}: The other six steps of the SA all provide graphical
front ends for the data required to populate the Semantic Robotic Description
Format (SRDF) and other configuration files used by MoveIt!. The SRDF provides
\textit{reconfigurable} semantic meta data of the robot model. It is data useful
to motion planning but not relevant to the URDF because it does not describe
physical properties of the robot. The SRDF information includes which set of
joints constitutes an arm and which set of links is considered part of the end
effector. It is one of the main components that allows MoveIt! to be robot
agnostic and to avoid dependencies on specific robots \cite{moveit}. Requiring
the user to configure all the semantic information by hand in a text editor
would be tedious and more difficult than using a GUI. The GUI populates the
available options for each input field in list boxes and guides the user through
filling in the necessary fields with buttons and graphical feedback.

The last step of the SA is to generate all launch scripts and configuration
files. This step outputs to file the information collected from the user during
the step-by-step user interface, as well as generates a series of default
configuration and launch scripts that are automatically customized for the
particular robot using the URDF and SRDF information. These defaults include
velocity and acceleration limits for each joint, kinematic solvers for each
planning group, available planning algorithms, and projection evaluators for
planning. Default planning adapters are setup for pre- and post-processing of
motion plans. Default benchmarking setups, controller and sensor manager
scripts, and empty object databases are all generated using launch scripts,
which essentially allow one to start different sets of MoveIt! functionality
that are already put together. 

These configuration files can easily be modified later from their default values by simply editing the text-based configuration files. The format of the files are based on ROS standards, which were chosen for their wide spread acceptance, readability, and simplicity. For the launch files an XML-based format custom to launching ROS applications was utilized. For all other configuration files the open source YAML data serialization format was used.

{\bf Automatic Runtime Tuning}: MoveIt! is designed to simplify solving planning
problems by reducing the number of hard-coded parameters and so called ``magic
numbers.'' Sampling-based planning algorithms in particular require such
parameters as input. MoveIt! uses heuristics from OMPL to automatically choose
good values for certain parameters to reduce the amount of expert domain
knowledge required and make MoveIt! extensible to a larger set of problems. 

An example of automatic runtime tuning is the resolution at which collision
checking is performed; it is defined as a fraction of the space extent. The
space extent is the lowest upper bound for the distance between the two farthest
configurations of the robot. This distance depends on the joint limits, the
types of joints, and the planning groups being used. Using the same information,
projections to Euclidean spaces can also be defined. These projections are used
to estimate coverage during planning. For example, the projections for robot
arms are orthogonal ones, using the joints closer to the shoulder, as those most
influence the position of the end-effector.

{\bf Benchmarking}: For applications that require more tuning and optimization than those afforded by automatically generated parameters and default values, MoveIt! provides the ability to configure and switch out different planning
components and and specify their configuration. However, this capability is much less useful
without the ability to quantify the results of different approaches.
Optimization criteria such as path length, planning time, smoothness, distance
to nearest obstacle, and energy minimization need benchmarking tools to enable
users and developers to find the best set of parameters and planning components
for any given robotic application.

MoveIt! lowers the barrier to entry to benchmarking by providing a command
line-based infrastructure and benchmarking configuration files that allows each
benchmark to easily be set up for comparison against other algorithms and
parameters \cite{cohen2012generic}. An additional GUI is currently in
development that makes benchmarking easier, more \textit{intuitive}, and reduces
the learning curve to this feature set of MoveIt!.

Choosing the best combination of planning components and parameters for any
particular robot and problem is a daunting task even for experts because of the
number of choices that must be made\cite{cohen2012generic}. A common method to
optimize an algorithms performance is to perform single and multivariable
parameter sweeps during benchmarking. MoveIt! provides an interface for this in
its benchmarking infrastructure by allowing an upper, lower, and increment
search values to be provided by the user. Results can be output into generic
formats for use in different plotting tools for analysis of which combination of
parameters performed the best.

Attempting to fine-tune the functionality of MoveIt! with benchmarking and parameter sweeping is a feature for expert users and it is generally not required for entry-level users.

\subsection{Easily Customize Framework Components}
\label{subsec:extensiblity}

MoveIt! lowers the barrier to entry by not requiring users to provide their own
implementation of any of the components in the motion planning framework. The
default planning components are based on OMPL, FCL, and KDL. However, these
default components are limiting to more advanced users who have their own
application or research-specific needs to fulfill. We will briefly describe here
how MoveIt! uses a plugin-based architecture and a high-level interface to
address these \textit{extensibility} issues (interested readers should refer to
\cite{moveit} for more detailed explanations). 

\begin{figure}[!t]
\centering
\includegraphics[width=3.4in]{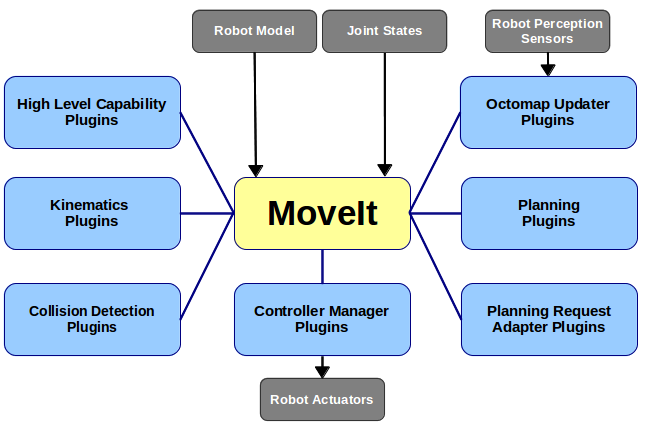}
\caption{Available planning component plugins for easily extending the
functionality of MoveIt!. Grey boxes represent external input and output.}
\label{fig:moveit_plugins}
\end{figure}

{\bf Plugins}: MoveIt! is designed to be \textit{extensible} by allowing its
various planning components to be customized through a lightweight plugin
interface \cite{moveit}. This is accomplished by using C++ shared objects that
are loaded at run time, reducing dependency complexities. This plugin-centric
framework, as seen in Figure \ref{fig:moveit_plugins}, provides interfaces for
forward and inverse kinematics, collision detection, planning, planning request
adapters, controllers, perception, and higher level capabilities. Almost
all aspects of MoveIt!'s functionality can be extended using plugins.

A particular strongpoint of MoveIt!'s feature set is its kinematics plugins that
it can automatically generate using the input URDF. The default KDL plugin uses
numerical techniques to convert from a Cartesian space to joint configuration
space. A faster solution can be achieved for some robots by utilizing OpenRave's
IKFast \cite{ikfast} plugin that analytically solves the inverse kinematics
problem. A combination of MoveIt! scripts and the IKFast Robot Kinematics
Compiler can automatically generate the C++ code and plugin needed to increase
the speed of motion planning solutions by up to three orders of magnitude
\cite{ikfast}.

Essentially, MoveIt! provides a set of data sharing and synchronization tools,
sharing between all planning components the robot's model and state. The
\textit{extensibility} of MoveIt!'s framework is greatly enhanced by not forcing
users to use any particular algorithmic approach. 

{\bf High Level Interfaces}: High level custom task scripting is easily
accomplished in MoveIt! with both a C++ and python interface that abstracts away
most of the underlying mechanisms in MoveIt!. Users who do not wish to concern
themselves with how the various low level planning components are operating can
focus instead on the high level application tasks, such as picking up an object
and manipulating it. Python in particular is a very easy scripting language that
enables powerful motion planning tasks to be accomplished with very little
effort.

\subsection{Documentation and Community}

Though common place in open source software projects \cite{bruyninckx2001open},
it should be mentioned for completeness that MoveIt! addressed the \textit{entry
barrier design principle} of \textit{documentation} by providing extensive
online wiki pages, a mailing list for questions, and a issue tracker for bug
reports and feature requests.

\section{Results}
\label{sec::results}

The success of MoveIt!'s efforts to lower its barrier of entry to new users
through the application of the barrier to entry principles is quantified in the
following. Its adoption rate, community activity, contributors, and results from
a user survey are used as indicators of its progress.

\subsection{Statistics}
\label{sec::statistics}

MoveIt! was officially alpha released on May 6th, 2013 --- about 11 months
prior to this writing. One method to quantify its popularity is by the total
number of binary and source code installations that have been performed. Though
not exactly representative of this data, MoveIt!'s website has an
``Installation'' page that receives an average of 940 unique page views per month
\cite{moveit} -- a large fraction of that number can be assumed to represent
unique installations. 

There are currently 312 members on the MoveIt! mailing list as shown over time in
Figure \ref{fig:membership_plot}. The posting activity of the mailing list over
time is also shown in Figure \ref{fig:membership_plot}, averaging 164 posts per
month since MoveIt! was launched.

\begin{figure}[!t]
\centering
\includegraphics[width=3.4in]{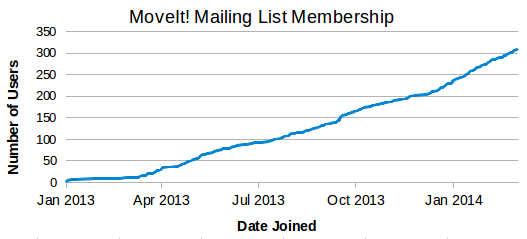}
\includegraphics[width=3.4in]{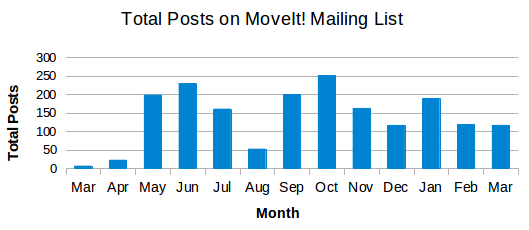}
\caption{MoveIt! mailing list statistics though March 2014. MoveIt! was alpha released in May 2013}
\label{fig:membership_plot}
\end{figure}

There have been a total of 63 contributors to the MoveIt! code base since its
initial development began in 2011. The total number of contributors over time is
shown in Figure \ref{fig:contributors}. According to statistics gathered by the
website Ohloh.com, which tracks activity of open source projects, MoveIt! is
``one the largest open-source teams in the world and is in the top 2\% of all
project teams on Ohloh'' \cite{ohloh}.

\begin{figure}[!t]
\centering
\includegraphics[width=3.4in]{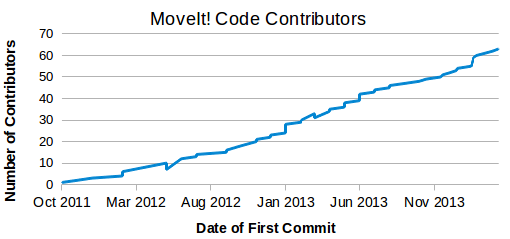}
\caption{MoveIt! source repository contributions through May 2014}
\label{fig:contributors}
\end{figure}

\subsection{Comparison}
\label{sec::comparison}

A brief comparison with MoveIt! to OMPL and OpenRave is shown in Figure \ref{fig:comparison}. In this diagram the total number of code contributors is plotted with respect to time, as reported from the projects respective version control system (VCS). No other software projects discussed in this paper had VCSs available publicly for comparison.

\begin{figure}[!t]
\centering
\includegraphics[width=3.4in]{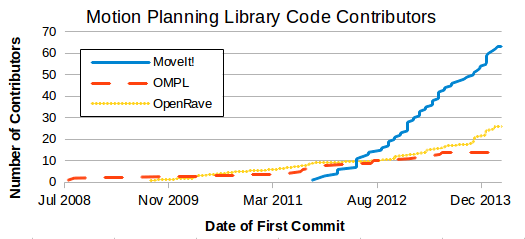}
\caption{Comparison of code contributions between MoveIt!, OMPL, and OpenRave}
\label{fig:comparison}
\end{figure}

\subsection{Survey}
\label{sec::survey}

\begin{figure}[!t]
\centering
\includegraphics[width=3.4in]{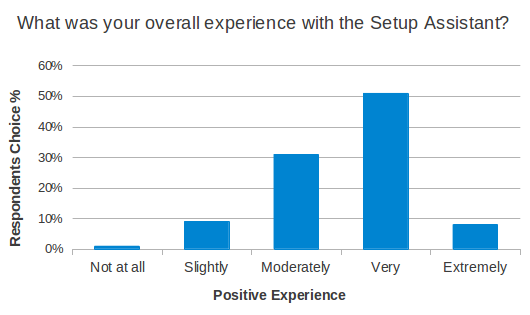}
\includegraphics[width=3.4in]{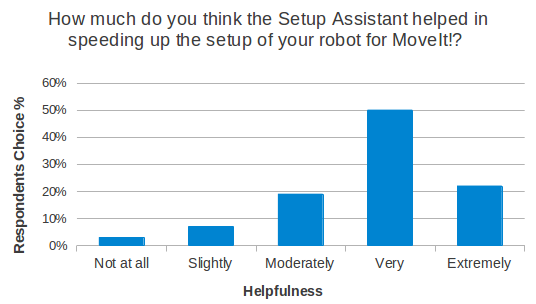}
\includegraphics[width=3.4in]{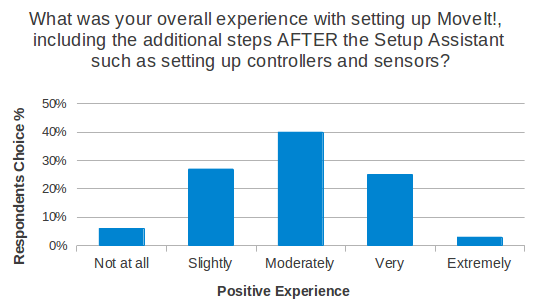}
\caption{Survey data of 105 respondents on the MoveIt! and ROS mailing lists.}
\label{fig:setup_assistant_positive}
\end{figure} 

A survey on users' experience with MoveIt! was administered on the MoveIt! and
ROS mailing lists. There were a total of 105 respondents; graduate students
represented by far the largest group of respondents (39\%), while
faculty/post-docs (18\%) and industry R\&D users (17\%) represented the next
biggest groups (see \cite{moveit} for the full survey results). Relevant results
corresponding to the use of the MoveIt! Setup Assistant are shown in Figure
\ref{fig:setup_assistant_positive}. 

Respondents were asked to rate their overall experience with using the MoveIt!
SA, and asked how much the SA helped in speeding up setup of a robot in MoveIt!.
For both questions, ninety percent had ``moderately'' to ``extremely'' positive
experiences with the SA, and for both questions over half rated their experience
as ``very good.''

Respondents then were asked what their overall experience was with setting up
and configuring MoveIt!, including any additional steps they had to take after
the SA, such as setting up controllers and sensors. In this question, the
results were less positive. Forty percent had a ``moderately'' positive
experience, but only 28\% had a ``very'' or ``extremely'' positive experience. 

An additional question asked respondents ``how many minutes would you estimate
you spent going from a URDF to solving motion plans using the MoveIt! Motion
Planning Visualizer?'' The responses to this question had a large amount of
variance, with the mean time taking users 1.5 hours with a standard deviation of
2 hours.

\section{Discussion}
\label{sec::discussion}

\subsection{MoveIt! Setup}
\label{sec::moveit_discussion}

We believe the barrier to entry for MoveIt! is easier than most, if not all,
open source motion planning software available today as discussed in Section
\ref{sec::existing}. As a result, MoveIt! has quickly become popular in the
robotics community as a powerful MPF that is extensible to most users' needs for
their robot application. The adoption rate of MoveIt! since its official release
half a year ago has been very positive in comparison to the size of the world
wide robotics community. With 309 users on the mailing list since MoveIt!'s
release, a new member has joined the project at a rate of nearly one per day.
These numbers indicate a healthy usage and popularity of this open source
software project.

Community effort to improve MoveIt! has been better than expected given the large number of code contributors during MoveIt!'s existence. The comparison of two other robotics software project's all time contributors in Figure \ref{fig:comparison} makes it very evident that MoveIt! is a popular robotics project relative to others. Ohloh's ranking of MoveIt! as one of the largest open source teams in the world confirms our belief that by making complex software more accessible, more developers will be able to report and fix issues. 

The results of the survey on MoveIt! indicated most people have found the
cornerstone of our approach to lowering the barrier of entry, the Setup
Assistant, to be very or extremely helpful in saving them time during setup
(72\% of respondents). Additionally, their overall experience was very or
extremely positive with the SA (59\%). However, in asking respondents their
overall configuration experience with MoveIt! beyond just the SA, their ratings
were lower, with only 28\% saying they had a very or extremely positive
experience setting up MoveIt!. This indicates that improvement can be made in
the overall integration process and that adding more steps and features to the
SA could reduce even further the entry barrier to MoveIt!. 

From the lower results from this last survey question, it is clear that the
setup and configuration process of MoveIt! can still be improved. A popular
response from some of the free-form questions in the survey is that setting up
the hardware controllers can also be a difficult task for non-experts, and the
MoveIt! setup process does not yet document and provide example code as well as
it could. It is likely this step will continue to require some custom coding to
account for arbitrary hardware interfaces and communication methods, but based
on the feedback we have received from actual users, this is certainly an area of
improvement for the MoveIt! Setup Assistant to address. 

The estimated setup time from taking a URDF and using MoveIt! to solve motion
plans shows a large range of variance and likely indicates that wide range of
experience levels in MoveIt! users. Although users averaged 1.5 hours to
configuring MoveIt! from scratch, 31\% reported it taking them 15 minutes or
less to setup MoveIt! for a new robot. Creating software powerful but simple
enough for all skill levels of users is a challenging task that MoveIt! will
continue to tackle.

Though not exactly within the scope of MoveIt!, creating the robot model itself
is a difficult task that typically requires a lot of trial and error in
configuring the links and joints properly. This process could be improved by a
better GUI for making arbitrary robot models, better tools for attaching the
links together correctly, and more documentation.

Finally, although MoveIt! is very extensible with its plugin-based architecture,
modifying the actual code base of MoveIt! can be intimidating due to its large
size. MoveIt! contains over 170 thousand lines of code across all its various
packages. Due to the need for computational speed and power, the layout of the
code can sometimes seem complicated and abstracted. 

We would like to emphasize the effect of a quick setup process and
\textit{Getting Started} demo on a new user unaccustomed to MoveIt! or motion
planning in general. The positive reinforcement of a quick initial success
encourages novices to continue to use the software and enables them to begin
going deeper into the functionality and code base. If the entry barrier is too
high, that is to say if it is too complex and error-prone, a new user will
likely give up and turn to other frameworks or custom solutions. Attempting to
blindly fix software that a new user has not had any success with is a very
difficult task.

\subsection{MoveIt Development}
\label{sec::moveit_development}

Finding the balance between the opposing objectives of the \textit{entry barrier design principles} was a difficult task in developing MoveIt!. \textit{Immediacy} was given one of the highest priorities, such that we focused on users being able to go from robot model to planning feasible motion plans with very few steps. To allow this, the GUI streamlined the entire process and only presented the most important and \textit{intuitive} configuration options. For example, the concept of defining the parts of a robot that make up an arm is very \textit{intuitive}. This focus on \textit{immediacy} sacrifices \textit{transparency} in that once users get this initial virtual demonstration working, they have not learned much on how to extend or dive deeper into the MPF. At this point \textit{documentation} is necessary to the user. 

Another pair of conflicting principles are \textit{extensibility} and \textit{intuitiveness}. The powerful plugin framework that MoveIt! provides allows custom components to be loaded and swapped out at runtime. However, this requires many layers of abstraction and inheritance, and results in overall convoluted code that is difficult for new developers. The balance is attempted by providing documentation and code examples for plugins that allows users to build new components without worrying about the underlying framework.

Integrating components from different sources, such as third party libraries of robotic software from other research groups, presents challenges as discussed in \cite{brugali2010component}. During MoveIt! development, the plugin interfaces were required to be general enough to work with many different implementation methods and choices of data structures. This was accomplished by providing ``wrapper'' packages that connect together the standard MoveIt! plugin interfaces with the third party software API. For example, MoveIt is currently setup to work with at least three planning libraries -- OMPL, SBPL \cite{likhachevsbpl}, and CHOMP \cite{ratliff2009chomp}. Although they represent fairly different approaches to motion planning and use different datastructures, each has a wrapper component that harmonizes them to work together in MoveIt!. It should be noted that as is true with any external dependency, maintaining compatibility with these wrappers has proven challenging.

\subsection{Robotic Software}
\label{sec::robotic_discussion}

The techniques utilized in lowering the barrier of entry for MoveIt! can easily
be applied to robotics software in general. Almost all robotics software
requires customizations specific to a particular hardware and kinematic
configuration. Reducing the difficulty of performing these customizations should
be the goal of robotic software engineers who desire to create useful tools for
a large audience. 

For example, perception applications such as visual servoing require similar
kinematic models to those being used in motion planning. Frame transforms must
be specified for the location of the camera and the location of the end effector
with respect to the rest of the robot's geometry \cite{visual_servoing}. This is
often a difficult and tedious task. Automating the setup and calibration of
these transforms lowers the barrier of entry to new users to robotic vision
software and makes the vision software useful to more users. In general,
automating the sequence of configuration steps necessary for performing
particular tasks is a useful strategy: the users will not have to think about
whether they have missed steps or whether they have performed the necessary
steps in the correct order. 

The \textit{entry barrier design principles} of immediacy, transparency, intuitiveness,
reconfigurability, extensibility, and documentation present a set of guidelines
for other open source robotic software projects to reduce their barriers of
entry to users. In fact, many existing robotics projects already follow subsets
of these principles, but typically to a lesser extent and fervor.

Creating a GUI such as the Setup Assistant is a time consuming process that many
robotics developers avoid in favor of hard-coded or command-line based
configuration, thereby neglecting the opportunity to attract non-expert users.
Between two developers, the various GUIs and configuration tools in MoveIt! took
about three months of development time. We believe that the trade-off in the
time invested is worthwhile for the higher adoption rates and creation of a
larger community willing to contribute to the software's development. 

It is understandable that when robotics research is the priority, spending time
on tangential aspects of the project such as GUIs and configuration tools can be
less important to the researcher. Still, we would like to encourage researchers
and developers alike, when possible, to spend the extra time making their work
reusable by taking into consideration the barriers to entry that other users
might encounter. Too often, software is touted as ``open source'' when its
usefulness is in actuality severely limited by the difficulties other users
encountered in applying it to other robotic projects. By sharing accessible open
source robotics software, the progress of robotic technology is accelerated and
the robotics community as a whole benefits.

\section{Conclusion}
\label{sec::conclusion}

Beyond the usual considerations in building successful robotics software, an
open source project that desires to maintain an active and large user base needs
to take into account the barriers of entry to new users. By making robotic
software more accessible, more users have the ability to utilize and contribute
to robotics development who previously could not have. The entry barrier design
principles are guidelines for robotic software engineers to improve the
usefulness and usability of their work to others as demonstrated in this paper
with the case study of MoveIt!.

As robotic algorithms become more complicated and the number of interacting
software components and size of the code base increases, configuring an
arbitrary robot to utilize robotic software becomes a daunting task requiring
domain-specific expertise in a very large breadth of theory and implementation.
To account for this, quick and easy initial configuration, with partially
automated optimization and easily extensible components for future
customization, is becoming a greater necessity in motion planning and in robotic
software engineering in general. 

\section*{Acknowledgments}
The authors would like to acknowledge E. Gil Jones as the initiator and Matthew
Klingensmith as the implementor of the Arm Navigation Setup Wizard, the
inspiration for the MoveIt! Setup Assistant.

\bibliographystyle{IEEEtran}
\bibliography{IEEEabrv,coleman_20131110}

\begin{IEEEbiography}[{coleman_20131110_f01}]{David Coleman}, M.S., 2013, is a
Ph.D. candidate in Computer Science at the University of Colorado. He obtained
his B.S. in Mechanical Engineering at Georgia Tech and Masters in C.S. at CU
Boulder. David's research interests include robotic manipulation, motion
planning, and controls. He is one of the main developers of MoveIt!, creating
the MoveIt! Setup Assistant while interning at Willow Garage continuing
development at the Open Source Robotics Foundation and with his regular
research.
\end{IEEEbiography}

\begin{IEEEbiography}[{coleman_20131110_f02}]{Ioan A. \c{S}ucan}, 
received the Ph.D. degree in computer science at Rice University,
Houston, TX, in 2011. His research interests include mobile
manipulation, task and motion planning for manipulation and motion
planning under differential constraints. While at Willow Garage
Ioan initiated the core components that make up the Arm
Navigation software platform, started the MoveIt! software platform
and the Open Motion Planning Library (OMPL). Ioan is the main
developer of MoveIt! and OMPL.
\end{IEEEbiography}

\begin{IEEEbiography}[{coleman_20131110_f03}]{Sachin Chitta}, Ph.D., is
associate director of robotics systems and software in the Robotics Program at
SRI International. His research interests include mobile manipulation, motion
planning and learning for manipulation. Sachin Chitta was at Willow Garage from
2007-2013 and was a core member of the team that developed the PR2 robot and the
Robot Operating System (ROS). He initiated and led the development of the
MoveIt! and Arm Navigation software platforms to enable advanced manipulation
capabilities for any robot. He obtained his PhD from the Grasp Lab at the
University of Pennsylvania in 2005.  
\end{IEEEbiography}

\begin{IEEEbiography}[{coleman_20131110_f04}]{Nikolaus Correll}, M.S., 2003,
Ph.D., 2007, is an Assistant Professor in Computer Science at the University of
Colorado. He obtained his Master's degree in Electrical Engineering from ETH
Z\"urich, a Ph.D. in Computer Science from EPFL and worked as a post-doc at MIT
CSAIL. Nikolaus's research interests are multi-robot and swarming systems. He is
the recipient of a 2012 NSF CAREER and a 2012 NASA Early Career Faculty
fellowship.
\end{IEEEbiography}



\vfill


\end{document}